\documentclass{article}

\PassOptionsToPackage{numbers, compress}{natbib}
    \usepackage[preprint]{neurips_2025}

\usepackage[utf8]{inputenc} % allow utf-8 input
\usepackage[T1]{fontenc}    % use 8-bit T1 fonts
\usepackage{hyperref}       % hyperlinks
\usepackage{url}            % simple URL typesetting
\usepackage{booktabs}       % professional-quality tables
\usepackage{amsfonts}       % blackboard math symbols
\usepackage{nicefrac}       % compact symbols for 1/2, etc.
\usepackage{microtype}      % microtypography
\usepackage{xcolor}         % colors
\usepackage{xpatch}
\usepackage{graphicx}
\usepackage{algorithm}
\usepackage{algpseudocode}
\usepackage{amsmath}
\usepackage{multirow}
\usepackage[table]{xcolor}
\usepackage{pifont}
\usepackage{tcolorbox}
\usepackage{listings}

\makeatletter
\xapptocmd{\NAT@bibsetnum}{\setlength{\leftmargin}{0pt}\setlength{\itemindent}{\labelwidth}\addtolength{\itemindent}{\labelsep}}{}{}
\makeatother

\title{DATE: \underline{D}ynamic \underline{A}bsolute \underline{T}ime \underline{E}nhancement for Long Video Understanding}

\author{
Chao Yuan$^{1,2}$, Yang Yang$^{4}$, Yehui Yang$^{3,\dagger}$, Zach Cheng$^{2}$ \\
\textsuperscript{1} Beihang University \quad 
\textsuperscript{2} Dcar, ByteDance \quad 
\textsuperscript{3} Qfin Holdings,Inc  \quad \\
\textsuperscript{4} MAIS, Institute of Automation, Chinese Academy of Sciences 
\\
{\tt\small yuanc3666@gmail.com, yang.yang@nlpr.ia.ac.cn} \\
{\tt\small yangyehuisw@126.com, chengyi.2024@bytedance.com} \\
}

\begin{document}

\maketitle

\begin{abstract}
Long video understanding remains a fundamental challenge for multimodal large language models (MLLMs), particularly in tasks requiring precise temporal reasoning and event localization. Existing approaches typically adopt uniform frame sampling and rely on implicit position encodings to model temporal order. However, these methods struggle with long-range dependencies, leading to critical information loss and degraded temporal comprehension. In this paper, we propose \textbf{D}ynamic \textbf{A}bsolute \textbf{T}ime \textbf{E}nhancement \textbf{(DATE)} that enhances temporal awareness in MLLMs through the Timestamp Injection Mechanism (TIM) and a semantically guided Temporal-Aware Similarity Sampling (TASS) strategy. Specifically, we interleave video frame embeddings with textual timestamp tokens to construct a continuous temporal reference system. We further reformulate the video sampling problem as a vision-language retrieval task and introduce a two-stage algorithm to ensure both semantic relevance and temporal coverage: enriching each query into a descriptive caption to better align with the vision feature, and sampling key event with a similarity-driven temporally regularized greedy strategy. Our method achieves remarkable improvements w.r.t. absolute time understanding and key event localization, resulting in state-of-the-art performance among 7B and 72B models on hour-long video benchmarks. Particularly, our 7B model even exceeds many 72B models on some benchmarks. 
\end{abstract}

\renewcommand{\thefootnote}{}
\footnotetext{$\dagger$ Project Leader. Codes: \textit{\url{https://github.com/yuanc3/DATE}}}

\begin{figure}[H]
\centering
\includegraphics[width=0.88\linewidth]{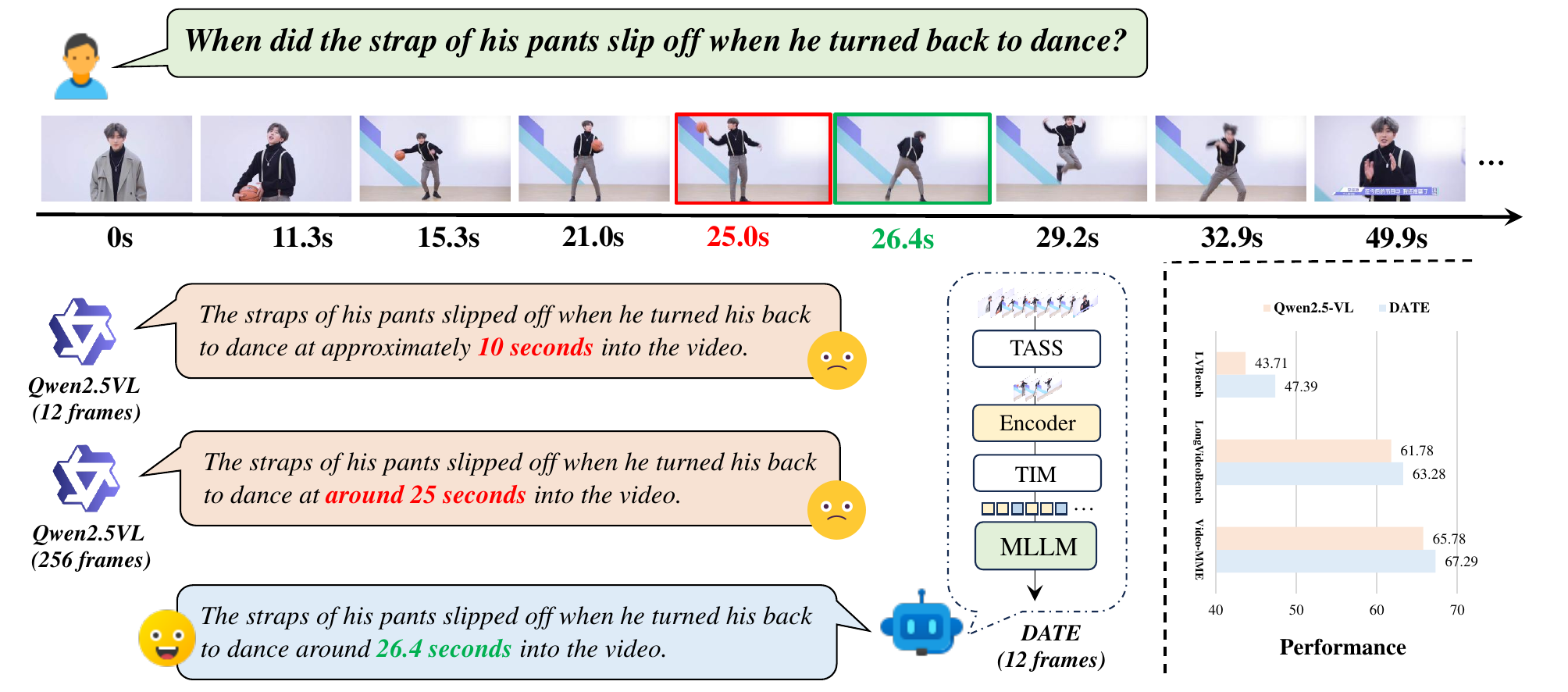}
\caption{A \textbf{Real} example of our proposed DATE compared with Qwen2.5-VL. It shows DATE with 12 frames beats 256 frames of Qwen2.5-VL.}
\label{fig:demo}
\end{figure}

% {%
% \begin{center}
% \begin{figure}
% \centering
% \includegraphics[width=\linewidth]{figs/demo.png}
% \caption{Two example of our proposed DATE compared with Qwen2.5-VL. The original video is 30 FPS. With our method, only 12 frames could beats 256 frames of Qwen2.5-VL.}
% \label{fig:demo}
% \end{figure}
% \end{center}
% % \begin{center}
% % \centering
% % \includegraphics[width=\linewidth]{figs/demo.png}
% % \caption{Two example of our proposed DATE compared with Qwen2.5-VL. The original video is 30 FPS. With our method, only 12 frames could beats 256 frames of Qwen2.5-VL.}
% % \end{center}%
% }

\section{Introduction}

Multimodal large language models (MLLMs)\cite{alayrac2022flamingo,cheng2024videollama,wang2024qwen2} have shown remarkable performance in a wide range of video understanding tasks, including video captioning, question answering, and event localization. However, when extended to long videos, these models face fundamental challenges in temporal reasoning and precise event localization. The essential reason for this limitation lies in the mismatch between rigid input length constraints of transformer architectures and the inherently long and continuous nature of real-world video content. As a result, existing approaches typically resort to uniform frame sampling as a preprocessing step. Unfortunately, this coarse-grained strategy often leads to the loss of critical visual events, temporal discontinuity, and the collapse of causality chains, severely limiting the model's capacity to reason over spatiotemporal structures. Moreover, there is no ability to perform perception and alignment of the absolute time and the corresponding frames.

One major obstacle is the inability of current methods to construct explicit representations of \textbf{absolute time}. Even when time-stamped subtitles are used as prompts, models struggle to align absolute timestamps with specific video frames. Although models such as Qwen2.5VL\cite{bai2025qwen2}  incorporate absolute time information into the temporal position embedding based on Multimodal RoPE\cite{wang2024qwen2,su2024roformer}, this approach exhibits critical drawbacks: For short video clips, time differences within one second remain indistinguishable; for long videos, the continual growth of positional indices leads to a loss of relative positional perception and eventual degradation of temporal comprehension. Our diagnostic experiments further confirm that such models do not solve problems related to absolute time reliably.

Another significant challenge comes from frame sampling itself. Uniform discretizations of frames lead to sparse observations, especially in long videos where adjacent frames may be separated by tens of seconds. Such sampling is agnostic to semantic content and fails to adapt dynamically to user queries, resulting in low recall when critical events are temporally sparse. Recent methods like Adaptive Keyframe Selection (AKS)\cite{tang2025adaptive} attempt to mitigate this by introducing query-guided dynamic sampling. However, they suffer from two key issues: (1) they use raw user questions as CLIP\cite{radford2021learning} text encoders, which contradicts CLIP's training paradigm centered on descriptive captions, leading to unstable or truncated representations; (2) their sampling method may still select irrelevant frames (e.g., negative samples with relatively high scores) and often fails in visually stable segments due to insufficient score variance.

To address these limitations, we proposed DATE, as shown in Fig.\ref{fig:framework}, for absolute time-aware video understanding and event localization. Our method builds a temporal coordinate system directly within the multimodal sequence by interleaving explicit timestamp tokens with video frame embeddings. This timestamp injection preserves visual continuity while allowing for precise and controllable temporal references. To guide the model towards relevant content, we formulate video sampling as a text-image retrieval task and employ a two-stage semantic-guided selection strategy: (i) rewriting user questions into caption-style descriptions for better alignment with CLIP-based vision-language similarity computation, and (ii) applying a temporally-regularized greedy sampling algorithm that ensures both high semantic relevance and temporal diversity. Our contributions are three-folds:

(1) We introduce \textbf{Timestamp Injection Mechanism (TIM)} that enables explicit absolute time modeling without modifying model weights or requiring additional training.

(2) We propose \textbf{Temporally-Aware Similarity Sampling (TASS)}, a temporally-regularized greedy sampling algorithm with semantic-guided caption generation to sample frames, which balance key events with video continuity.

(3) We show that our method achieves superior \textbf{ spatial perception} and \textbf{event localization}, especially for \textbf{hour-long} video scenarios, which achieve SOTA on 7B models, even surpassing many 72B models. Moreover, the DATE-72B model achieves state-of-the-art performance.

\section{Related Works}

\subsection{Multimodal Large Language Models for Video Understanding}

With the widespread success of large language models (LLMs) \cite{achiam2023gpt,brown2020language,chiang2023vicuna,chowdhery2023palm,chung2024scaling,grattafiori2024llama,touvron2023llama,touvron2023llama2,ray2023chatgpt,chen2024fewer} in natural language processing, researchers have extended these models to multimodal scenarios, forming multimodal large language models (MLLMs)\cite{lai2024lisa,liu2023visual}. By incorporating visual encoders, MLLMs are capable of processing visual inputs such as images or videos, enabling tasks like visual question answering, video captioning, and visual reasoning\cite{maaz2023video,alayrac2022flamingo,chen2024sharegpt4video,wu2024dibs,min2024morevqa,qian2024momentor,wang2022negative}. Representative models include Video-ChatGPT\cite{maaz2023video,lin2023video}, LLaVA-Video\cite{zhang2024video}, VideoLLAMA\cite{zhang2023video,cheng2024videollama,zhang2025videollama}, and Qwen-VL\cite{wang2024qwen2,bai2025qwen2}, which typically encode video frames into visual tokens and feed them into the model alongside textual tokens. However, due to the inherent context length limitations of LLMs, these models often rely on fixed frame sampling strategies, resulting in significant information compression when processing long video data\cite{fu2024videomme,wu2024longvideobench,wang2024lvbench}. Moreover, long videos present unique challenges such as sparse events and wide semantic spans, which demand more effective temporal modeling and cross-segment reasoning capabilities. Therefore, many strategies \cite{shang2024interpolating,zhang2024long,wei2024visual,chen2024longvila,wang2025adaretake,cheng2024enhancing,he2024zipvl,he2024ma} proposed for longer context.

\begin{figure}
\centering
\includegraphics[width=\linewidth]{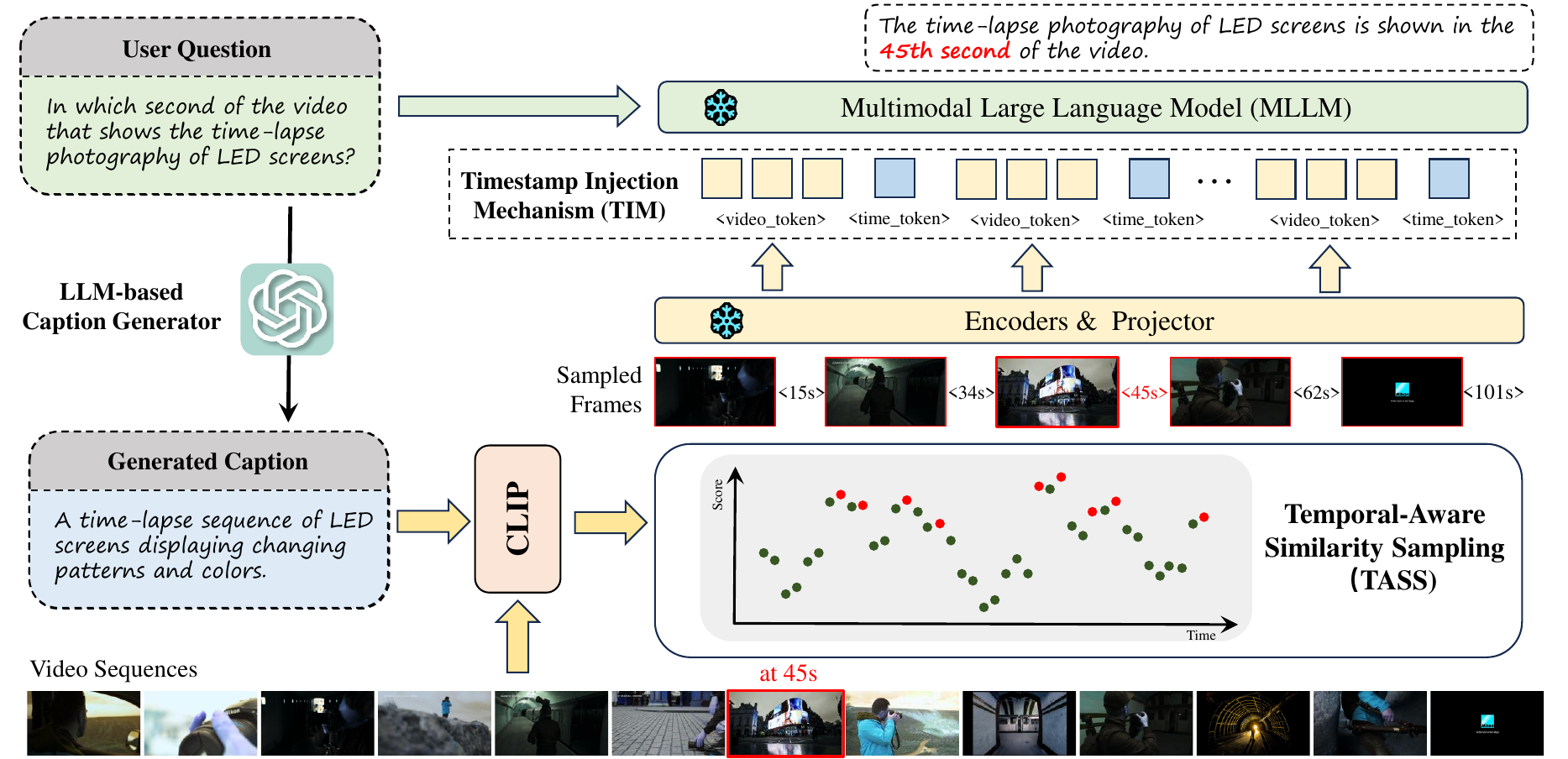}
\caption{Overview of the proposed framework. For each user input question, using LLM-based Caption Generator to generate a CLIP-aligned image caption, and calculate the similarity with video frames. Then, use Temporal-Aware Similarity Sampling (TASS) strategy to sample the frames (The real sampled frames and orders of this demo could be found in \textbf{Appendix B}). Last, with Timestamp Injection Mechanism (TIM), we embed timestamps aligned with each frame.}
\label{fig:framework}
\end{figure}

\subsection{Temporal Modeling}

Temporal modeling is a fundamental challenge in long video understanding. Existing methods can be broadly categorized into two groups: \ding{172}Using data with timestamps to fine-tune model with time tokens\cite{chen2024timemarker} or prompts with timestamps\cite{ren2024timechat}. These need more data and training cost. \ding{173}Explicit incorporation of time into positional encoding. For example, Qwen2.5VL introduces MRoPE\cite{bai2025qwen2} and Qwen2.5-Omni\cite{xu2025qwen2} introduces TMRoPE, which use absolute time signals into its rotary positional encoding. However, this encoding mechanism is prone to positional drift in long sequences, where the encoded position values grow too quickly with sequence length, thereby distorting the relative temporal relationships between frames. This can reduce the ability of the model to capture temporal causality and duration. More importantly, these methods often fail to provide a stable temporal reference, thus limiting the ability of the model to perceive absolute time.

% To enhance the model’s temporal awareness, we propose a simple yet effective timestamp embedding strategy. By inserting an explicit time token after each video token. This creates a continuous and interpretable time axis without modifying the model architecture. Such a structure serves as a strong temporal anchor, encouraging the model to associate specific visual content with distinct time points. It also facilitates temporal segmentation and improves contextual understanding across video segments—especially in tasks that require fine-grained temporal localization.

\subsection{Frame Sampling Strategy}
To mitigate the performance bottleneck caused by limited input length, frame sampling has become a crucial component in video understanding systems. The most common strategy is uniform sampling\cite{bai2025qwen2,cheng2024videollama,li2024llavaonevision}, which is straightforward but fails to adaptively select frames based on semantic importance. This often leads to omission of critical content, especially in videos with dense or uneven event distributions. To address this, some semantics-aware frame selection methods with VLMs like CLIP\cite{radford2021learning} have been proposed, such as BOLT\cite{liu2025bolt} and AKS\cite{tang2025adaptive}, and they proved to be effective over uniform and topk sampling. However, they all use question to find frames, this is not a good way for CLIP to embed question, since it was not trained with question. Meanwhile, they may also sample negative frames and loss critical temporal continuity (action, movement, etc.).

\section{Methods}

\begin{figure}
\centering
\includegraphics[width=0.85\linewidth]{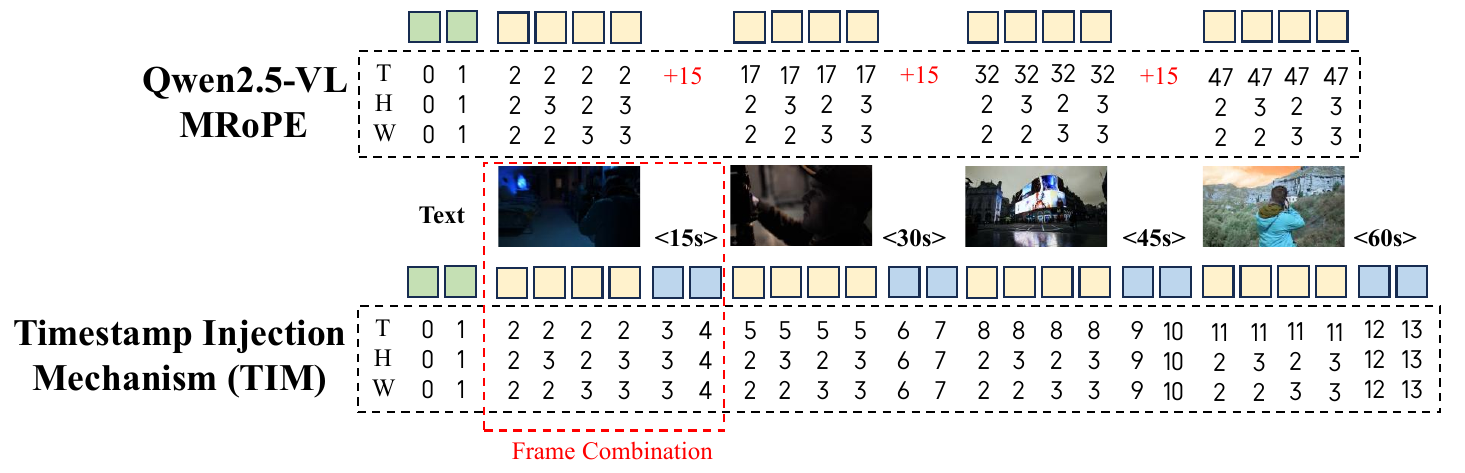}
\caption{The Multimodal RoPE (MRoPE) with our Timestamp Injection Mechanism (TIM) compared with Qwen2.5-VL's MRoPE. \textbf{Qwen2.5-VL: } Add 15 since there are 15 seconds betweet frames. \textbf{TIM(ours): }The temporal dimension $T$ is extended with time token. The spatial dimensions ($H, W$) remain aligned with the first frame, ensuring spatial consistency across the whole sequence.}
\label{fig:tim}
\end{figure}

\subsection{Timestamp Injection Mechanism (TIM)}

To enhance the temporal perception of Multimodal Large Language Models (MLLMs) in video understanding, especially in long videos requiring absolute time localization, we propose a timestamp injection mechanism. This mechanism is model-agnostic and compatible with most mainstream MLLMs. In this work, we take Qwen2.5-VL\cite{bai2025qwen2}, which incorporates explicit absolute time encoding, as our baseline method.

\paragraph{Token-Level Timestamp Injection}

The latest open-source MLLM, Qwen2.5-VL, relies on their proposed MRoPE (Multimodal RoPE) mechanism to model temporal sequences with time interval in the position ID of MRoPE\cite{wang2024qwen2}, to embed absolute time of video frames. However, our experiments demonstrate that this approach lacks a true understanding of absolute time.

To address this, we introduce a token-level timestamp injection mechanism. As shown in Fig.\ref{fig:tim}, for each sampled frame, we construct the input sequence using an interleaved structure of visual and time tokens:
\[
\small
\texttt{\textcolor{red}{<video\_token><time\_token>}\textcolor{green}{<video\_token><time\_token>} \dots \textcolor{blue}{<video\_token><time\_token>}}
\]

Here, each color represents the combination of video tokens and timestamps of a frame, \texttt{<video\_token>} represents the visual tokens (not one token), and \texttt{<time\_token>} is its corresponding textual timestamp (e.g., \texttt{01:23} or \texttt{83s}). This structure preserves visual continuity while injecting a precise and controllable temporal reference, enabling the language model to perform time-aware reasoning task such as event ordering and absolute time localization.

\paragraph{Reconstruction of Positional Encoding and Sequential Normalization}

The MRoPE mechanism in Qwen2.5-VL introduces absolute time information via position indices in the visual branch. Although it models temporal order to some extent, it suffers from critical limitations when applied to long videos due to linearly increasing position indices(IDs):

\textbf{(1) }\textbf{Sparsity and Resource Inefficiency:} Since position IDs grow proportionally, large time gaps (e.g., 20s between frames) leading to inefficient use of the sequence length and potential index explosion (e.g., 10,000 in hour-long videos).

\textbf{(2)} \textbf{Degradation of Relative Positional Awareness:} Large gaps between position IDs disrupt the relative distances between tokens, compromising the ability to capture local temporal structures.

To mitigate these issues, we remove the absolute time component from Qwen2.5VL's MRoPE and retain only the original Multimodal RoPE (MRoPE) encoding. Specifically, the temporal dimension $T$ is encoded using a simple \textit{sequential indexing} strategy, where position indices increment according to the natural order of tokens. Furthermore, to preserve the spatial encodings between video frames, we ensure that only the temporal dimension $T$ is extended along with time token insertion. The spatial encodings ($H, W$) remain aligned with the first frame, ensuring spatial consistency across the sequence.

This design maintains the numerical stability of RoPE\cite{su2024roformer}, and preserves the model's sensitivity to token order. Meanwhile, absolute time perception is handled independently via the explicit \texttt{<time\_token>}s, resulting in a decoupled and robust time representation framework.

\subsection{Temporal-Aware Similarity Sampling (TASS)}

Discretized video frame sampling is a common preprocessing step in multimodal video modeling. However, in long video scenarios, uniformly spaced sampling strategies exhibit clear limitations. On the one hand, the temporal gaps between frames may span several seconds to minutes, making it likely to miss sparse but semantically critical moments. On the other hand, uniform sampling is task-agnostic, severely undermining the recall of key events.

Sampling directly based on similarity leads to frames with little variation being sampled continuously, which results in video features collapsing into a single image. Sampling across too large a span would then lead to problems with key event continuity, difficulty in recognizing object movement, etc., i.e., a similar problem to that which would occur with uniform sampling and AKS\cite{tang2025adaptive}. 

Thus, we proposed \textbf{TASS}, a temporally-regularized greedy sampling algorithm that ensures both high key event continues and temporal diversity. It consists of two main stages: (i) \textit{semantic-enhanced similarity computation}, and (ii) \textit{similarity-prioritized sampling under temporal constraints}. 

\paragraph{Semantic Enhancement: From Question to Caption}

To improve the consistency of the visual-language alignment, we first convert the user's query (typically a question) into a more descriptive caption using a language model, and the prompt of this step can be seen in Appendix \ref{appendix:e}. Unlike raw questions, captions exhibit a declarative style that aligns better with CLIP's image-text matching paradigm, activating more stable and complete semantic representations.

Each video frame \( v_i \) is embedded using CLIP, and its similarity to the caption \( c \) is calculated as:
\begin{equation}
s_i = \text{CLIP}(v_i, c) = \frac{ \langle v_i, c \rangle }{ \|v_i\| \cdot \|c\| }
\end{equation}

\paragraph{Temporal-Aware Similarity Sampling}

We first compute a dynamic threshold $s_{\text{mean}}$ which is the mean of all similarity scores. Scores below the mean are considered \textit{negative samples}, as they contribute little to answering the user's query and are therefore discarded. To ensure computational efficiency, we further cap the number of top-ranked candidates by setting an upper bound proportional to the final number of selected frames, i.e., \( \text{topk} \leq 4 \times \text{max\_frames} \).
\begin{equation}
\text{topk} = \min\left( \left|\left\{ i \mid s_i > s_{\text{mean}} \right\}\right|,\, \alpha \times \text{max\_frames} \right)
\end{equation}
where $\alpha$ is a controllable coefficient. It denotes the number of frames to be sampled (candidate frames). For example, Qwen2.5-VL-7B can process up to 256 frames, and we set $\alpha=4$ by default, using our sampling strategy, we can effectively compress and select representative frames from a sequence of $4*256=1024$ frames. When negative sample filtering is considered, the expected number of candidate frames for sampling could be 2048.

While many continuous frames are semantically aligned, they often cluster temporally, leading to redundancy. To ensure temporal diversity while preserving semantic relevance, we introduce a greedy selection algorithm that is similarity first with enforcing a minimum time interval \( \delta \) between selected timestamps. If fewer than \(N_{\text{max}}\) frames are obtained, \(\delta\) is iteratively decayed until the quota is met. The pseudo-code is as follows:

\begin{algorithm}[H]
\small
\caption{Temporal-Aware Similarity Sampling (TASS)}
\label{alg:tass}
\begin{algorithmic}[1]
\Require Top-K timestamps \( \mathcal{I}_{\text{topK}} \), sampled frames \( N_{\text{max}} \), initial interval \( \delta_0 \)
\Ensure Selected timestamps \( \mathcal{S}_{t} \)
\State Initialize \( \mathcal{S}_{t} \gets \emptyset \), \( \delta \gets \delta_0 \), decay ratio \( \lambda = 0.5 \)
\While{ \( |\mathcal{S}_{t}| < N_{\text{max}} \) }
  \For{each \( t_k \in \mathcal{I}_{\text{topK}} \) }
    \If{ \( \forall t_j \in \mathcal{S}_{t}, |t_k - t_j| \geq \delta \) or \( \mathcal{S}_{t} = \emptyset\) }
      \State \( \mathcal{S}_{t} \gets \mathcal{S}_{t} \cup \{t_k\} \)
      \State Remove \( t_k \) from \( \mathcal{I}_{\text{topK}} \) 
      \If{ \( |\mathcal{S}_{t}| \geq N_{\text{max}} \) }
        \State \textbf{break}
      \EndIf
    \EndIf
  \EndFor
  \State \( \delta \gets \delta \cdot \lambda \)
\EndWhile
\State \Return sorted \( \mathcal{S}_{t} \)
\end{algorithmic}
\end{algorithm}

The most relevant work w.r.t. TASS is the Adaptive Keyframe Selection (AKS) proposed by Tang et al.\cite{tang2025adaptive}, which introduces a query-driven sampling mechanism. However, it suffers from two major issues: (1) It directly uses raw questions as CLIP text inputs, misaligned with CLIP’s caption-style since it was trained with image-caption pairs but not questions, and prone to semantic truncation due to the input limitation; (2) Its variance-based sampling strategy tends to include false positives (i.e., high-scoring frames from negative segments), due to the small magnitude of score variations, and may miss keyframes in visually smooth regions.

In contrast, our method leverages caption rewriting for better alignment and introduces a temporal regularization mechanism to ensure broader temporal coverage. This makes sampling more robust and effective for modeling temporally distributed events in long videos.

\section{Experiments}

\subsection{Benchmarks}
To comprehensively evaluate our proposed \textsc{DATE} on long video understanding, we conduct experiments on three hour-long video benchmarks that emphasize complex temporal reasoning and long-context modeling:

\textbf{Video-MME}\cite{fu2024videomme} is a multimodal evaluation benchmark designed for general video understanding. It contains 900 videos (256 hours in total) across various categories and durations, annotated with 2,700 expert-curated multiple-choice QA pairs. The dataset is partitioned into short (<2 min), medium (4–15 min), and long (30–60 min) subsets, enabling a detailed analysis of temporal scalability.

\textbf{LongVideoBench}\cite{wu2024longvideobench} focuses on long-context multimodal reasoning. It comprises 3,763 videos of up to 1 hour in length and 6,678 annotated questions across 17 categories. The benchmark emphasizes fine-grained temporal retrieval and localized event reasoning, making it ideal for evaluating absolute time comprehension.

\textbf{LVBench}\cite{wang2024lvbench} is one of the most challenging benchmarks for long video understanding, with an average video length of over 4,000 seconds. It provides 1,549 QA pairs including multiple tasks such as entity tracking, temporal grounding, and causal reasoning, offering a comprehensive testbed for temporal-aware video modeling.

\textbf{Implementation Details}
We adopt Qwen2.5-VL (7B and 72B)\cite{bai2025qwen2} as our baseline model. For fair comparison and reproducibility, we utilize the publicly released checkpoints and re-evaluated all benchmarks following their official technical report. Our DATE also follows the same settings. In the evaluation, the baseline adopts a uniform sampling rate of 4 FPS, with the resolution set to 448 (longest side) and a maximum of 256 input frames across all benchmarks. All the experiments are conducted with Nvidia A100-80G GPUs.
For our proposed TASS, deepseek-v3\cite{liu2024deepseek} is used for caption generation. Then, the frames are extracted with 1 FPS for all videos to calculate the visual-textual similarity score with the generated caption. Visual-textual similarity is computed using the CLIP ViT-B/32\cite{radford2021learning} model to enable the semantic-aware frame filtering. In the TASS (Temporal-Aware Similarity Sampling) module, we set the selection ratio coefficient $\alpha=4$, and initialize the temporal interval constraint $\delta_0$ to 20 seconds.

\begin{table}[]
\caption{Performance comparison on long video benchmark with SOTAs, including Video-MME (w/o subtitles), LongVideoBench, and LVBench. For fairly comparison, we re-test the model based on the technical report disclosed by QwenVL team, with all video inputs preprocessed based on 4FPS and 448 resolution. (\ding{171}: official reported results. \ding{168}: we re-test results). In the test, we found that the metric reported by QwenVL team on LongVideoBench were tested at 224 resolution. }
\small
\begin{tabular}{ccccccc}
\hline
\multirow{2}{*}{Models}&\multirow{2}{*}{Size}&\multirow{2}{*}{Frames}&\multicolumn{2}{c}{Video-MME (w/o sub)}&$\text{LongVideoB}$&$\text{LVBench}$\\\cline{4-7}
&&&\begin{tabular}[c]{@{}c@{}}Long\\(30-60min)\end{tabular}&\begin{tabular}[c]{@{}c@{}}Overall\\(0-60m)\end{tabular}&\begin{tabular}[c]{@{}c@{}}val\\(8s-3600s)\end{tabular}&\begin{tabular}[c]{@{}c@{}}val\\(avg.\textgreater{}4000s)\end{tabular}\\\hline
\multicolumn{7}{c}{\textit{Closed Video MLLMs}}\\\hline
\textcolor{gray}{GLM-4V-Plus}&\textcolor{gray}{-}&\textcolor{gray}{256}&\textcolor{gray}{-}&\textcolor{gray}{70.8}&\textcolor{gray}{-}&\textcolor{gray}{58.7}\\
\textcolor{gray}{GPT-4o}&\textcolor{gray}{-}&\textcolor{gray}{384}&\textcolor{gray}{65.3}&\textcolor{gray}{71.9}&\textcolor{gray}{66.7}&\textcolor{gray}{27}\\
\textcolor{gray}{Gemini-1.5-Pro}&\textcolor{gray}{-}&\textcolor{gray}{1/0.5fps}&\textcolor{gray}{67.4}&\textcolor{gray}{75}&\textcolor{gray}{64}&\textcolor{gray}{33.1}\\\hline
\multicolumn{7}{c}{\textit{Open-source Video MLLMs\textgreater{}70B}}\\\hline
% \textcolor{gray}{LLaVA-OneVision-72B\cite{li2024llavaonevision}}&\textcolor{gray}{72B}&\textcolor{gray}{32}&\textcolor{gray}{-}&\textcolor{gray}{66.2}&\textcolor{gray}{61.3}&\textcolor{gray}{-}\\
% \textcolor{gray}{Qwen2.5-VL-72B\cite{bai2025qwen2}}&\textcolor{gray}{72B}&\textcolor{gray}{768}&\textcolor{gray}{-}&\textcolor{gray}{73.3}&\textcolor{gray}{60.7}&\textcolor{gray}{47.3}\\
% \textcolor{gray}{InternVL2.5-78B\cite{chen2024intervl25}}&\textcolor{gray}{78B}&\textcolor{gray}{16-64}&\textcolor{gray}{-}&\textcolor{gray}{72.1}&\textcolor{gray}{63.6}&\textcolor{gray}{-}\\
% \textcolor{gray}{InternVL3-78B\cite{zhu2025internvl3}}&\textcolor{gray}{78B}&\textcolor{gray}{16-64}&\textcolor{gray}{-}&\textcolor{gray}{72.7}&\textcolor{gray}{65.7}&\textcolor{gray}{-}\\\hline
LLaVA-OneVision-72B&72B&32&-&66.2&61.3&-\\
LLaVA-Video&72B&64&61.5 &70.6 & 61.9&-\\
Qwen2-VL&72B& 768&62.2 & 71.2 &60.4 &41.3\\
InternVL2.5-78B&78B&16-64&-&72.1&63.6&-\\
InternVL3-78B&78B&16-64&-&72.7&65.7&-\\
% Qwen2.5-VL-72B\cite{bai2025qwen2}&72B&768&-&73.3&60.7&47.3\\
\rowcolor{gray!15}
Qwen2.5-VL-72B\ding{171}&72B&768&-&73.3&60.7&47.3\\
\rowcolor{gray!15}
Qwen2.5-VL-72B\ding{168}&72B&256&63.4&72.7&66.9&48.8\\
\rowcolor{gray!15}
\textbf{DATE-72B(Ours)}&72B&\textbf{256}&\textbf{65.3}&\textbf{73.3}&\textbf{68.1}&\textbf{52.1}\\
\hline
\multicolumn{7}{c}{\textit{Small Video MLLMs}}\\\hline
VITA-1.5&7B&16&47.1&56.1&-&-\\
LLaVA-Video&7B&64&-&63.3&58.2&-\\
NVILA&8B&256&54.8&64.2&57.7&-\\
ByteVideoLLM&14B&256&56.4&64.6&-&-\\
VideoLLaMA3&7B&180&-&66.2&59.8&45.3\\
InternVL3-8B&8B&16-64&-&66.3&62.5&-\\\hline
\rowcolor{gray!15}
Qwen2.5-VL-7B\ding{171}&7B&256&-&65.1&$\text{56.0}_{224dpi}$&45.3\\
\rowcolor{gray!15}
Qwen2.5-VL-7B\ding{168}&7B&256&55.4&65.8&$\text{61.8}_{448dpi}$&43.7\\
\rowcolor{gray!15}
\textbf{DATE-7B(Ours)}&7B&256&\textbf{57.3}&\textbf{67.3}&\textbf{63.3}&\textbf{47.4}\\\hline
% \rowcolor{gray!15}
% \textcolor{gray}{Qwen2.5-VL-72B\cite{bai2025qwen2}}\ding{171}&\textcolor{gray}{72B}&\textcolor{gray}{768}&\textcolor{gray}{-}&\textcolor{gray}{65.1}&\textcolor{gray}{56.0}&\textcolor{gray}{45.3}\\
% \rowcolor{gray!15}
% \textbf{DATE-72B(Ours)}&72B&256&\textbf{61.3}&\textbf{}&\textbf{68.1}&\textbf{52.1}\\\hline
\end{tabular}
\label{tab:main}
\end{table}

\begin{figure}
\centering
\includegraphics[width=0.75\linewidth]{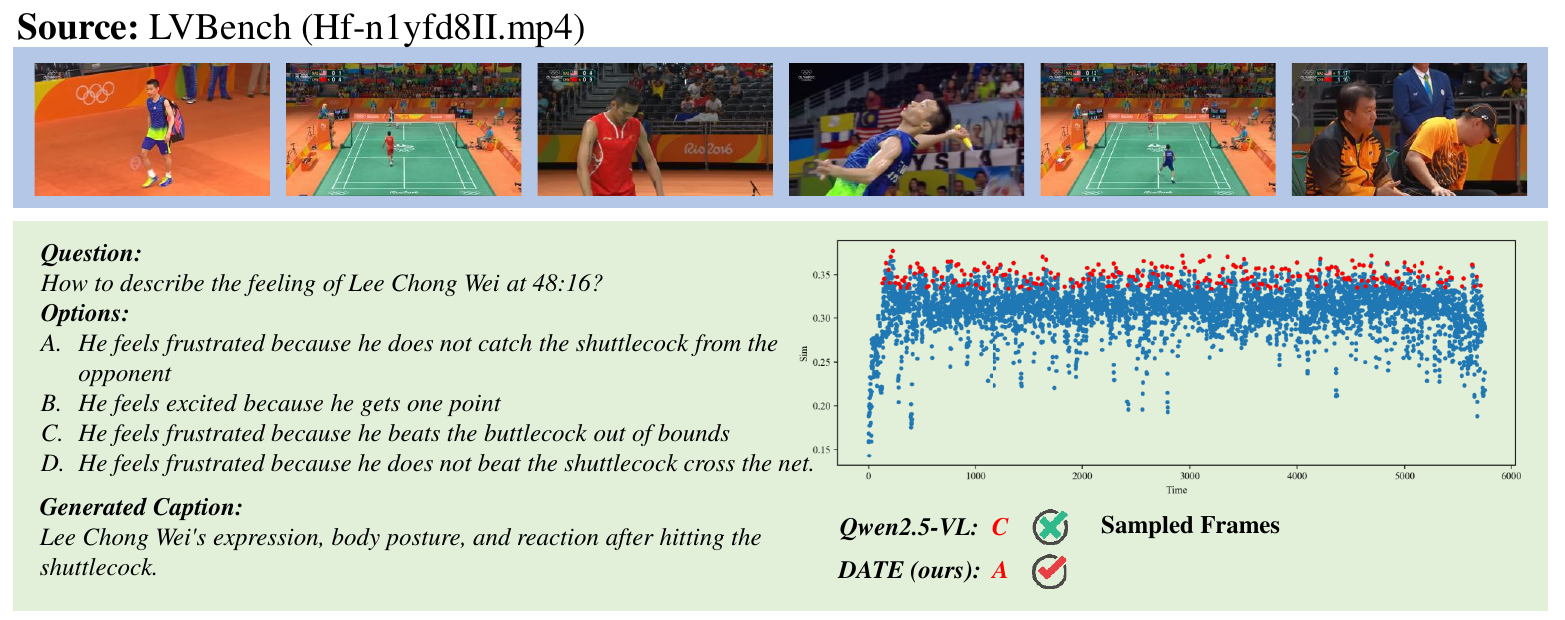}
\caption{A real demo compared DATE-7B with Qwen2.5-VL-7B. The caption is generated with our method and calculate similiarity scores with frames. The \textcolor{red}{\textbf{red}} points are sampled frames with TASS. More could be foung in \textbf{Appendix}.}
\label{fig:main_demo}
\end{figure}
% \begin{figure}
% \centering
% \includegraphics[width=\linewidth]{figs/pdfs/demo.pdf}
% \caption{Two examples of our proposed DATE compared with Qwen2.5-VL. The original video is 30 FPS. With our method, only 12 frames could beats 256 frames of Qwen2.5-VL.}
% \label{fig:demo}
% \end{figure}

\subsection{Main Results}

\paragraph{Comparison with the State-of-the-Art}
We compare our proposed method, DATE, with a variety of state-of-the-art closed-source and open-source video MLLMs on multiple long-video benchmarks, as summarized in Table~\ref{tab:main}. Compared to other small-scale video MLLMs, DATE achieves consistent improvements across all benchmarks, outperforming the prior best model (Qwen2.5-VL) by +1.5\% on Video-MME (Overall), +1.5\% on LongVideoBench (val), and +2.1\% on LVBench (An extremely long video benchmark). Moreover, our method (256 frames) even outperforms the Qwen2.5-VL-72B (768 frames) model on LongVideoBench and LVBench. These gains demonstrate DATE's superior temporal modeling capability, especially in handling extremely long videos. It shows our methods effectively injects temporal cues and helps the model focus on semantically important moments, enabling more robust long-range reasoning.

\begin{figure}[]
\centering
\begin{minipage}{0.48\textwidth}
    \centering
    \includegraphics[width=\textwidth]{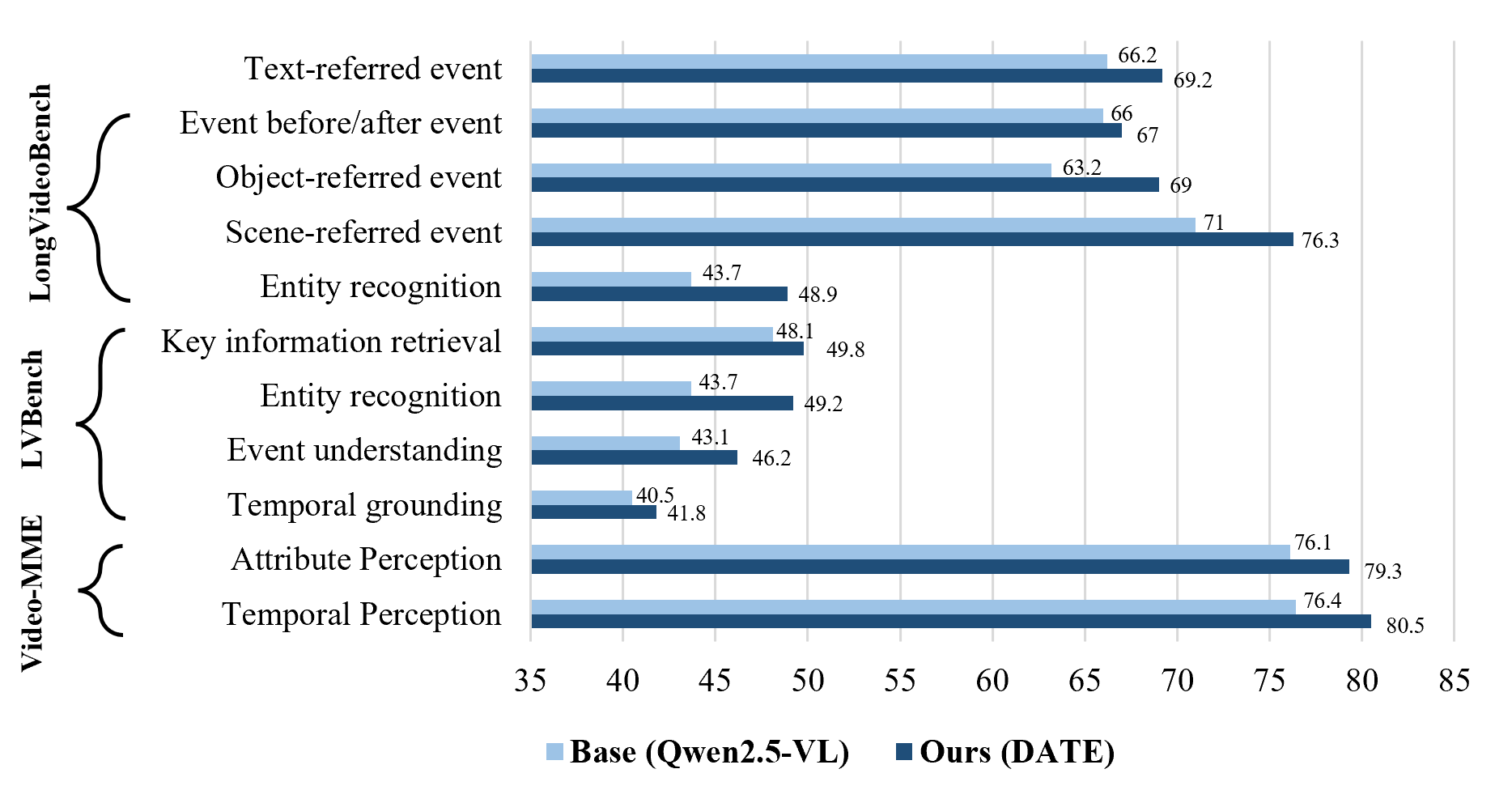}
    \caption{Comparison of performance related to event-aware tasks in the three benchmarks: Video-MME, LongVideoBench, and LVBench.}
    \label{fig:event}
\end{minipage}
\hfill
\begin{minipage}{0.48\textwidth}
\centering
\begin{table} [H]
    \caption{Ablation study on two components of DATE-7B on three long video benchmarks: Video-MME, LongVideoBench, and LVBench.}
    \footnotesize
    \begin{tabular}{cc|ccc}
    \hline
    TIM & TASS  & V-MME & $\text{LongVideoB}$ & $\text{LVB}$ \\ \hline
     \ding{55} & \ding{55}   & 65.8&	61.8	&43.7  \\
     \ding{51} & \ding{55}    &66.5&	61.9&	44.9 \\
    \ding{55} & \ding{51}    & 66.6&	62.8&	46.7 \\
    \rowcolor{gray!15}
    \ding{51} & \ding{51}    & \textbf{67.3}&	\textbf{63.3}&	\textbf{47.4 }\\ 
    \hline
    \end{tabular}
    \label{tab:ablation}
\end{table}
\end{minipage}
\end{figure}

\paragraph{Comparison with Event-aware tasks.}
To better understand the advantage of DATE in modeling temporal and event-centric information, we provide a detailed comparison across fine-grained sub-tasks in Video-MME, LVBench, and LongVideoBench, as shown in Figure~\ref{fig:event}.

% Compared to the Qwen2.5-VL, our method yields consistent gains across all event-aware tasks. In Video-MME, DATE improves both Temporal Perception and Attribute Perception by +4.1\% and +3.2\%, respectively. In LVBench, DATE particularly excels at Temporal Grounding (+2.7\%) and Event Understanding (+2.5\%), indicating enhanced temporal alignment and semantic abstraction capabilities.

% In LongVideoBench, which focuses on fine-grained reasoning over long temporal spans, DATE achieves notable improvements in all event-centric metrics: +5.3\% on Scene-referred events, +5.8\% on Object-referred events, +1.0\% on Event before/after event, and +3.0\% on Text-referred events. These results validate DATE’s ability to better encode temporal relations and contextual semantics, especially under challenging long-duration scenarios. We hypothesize that the introduction of timestamp tokens and keyframe-aware modeling enables more effective temporal referencing and memory localization, leading to better performance in event-centric understanding tasks.

\subsection{Precise event localization capabilities}

Our DATE shows significant advantages in accurate event localization. As shown in the Fig.\ref{fig:demo}, DATE can accurately identify the specific time points of events even when only 12 frames are used, and even accurately samples the critical time with only one frame as shown by the sampling order labeled in the sampling graph. However, the baseline model still shows significant deviations at 256 frames. This validates the effectiveness and robustness of our proposed temporal modeling and semantic-driven sampling strategy for long video understanding. Fig.\ref{fig:main_demo} also shows some cases in benchmarks, more examples can be found in the \textbf{Appendix}.

% \begin{figure}
% \centering
% \includegraphics[width=\linewidth]{figs/demo.png}
% \caption{Two example of our proposed DATE compared with Qwen2.5-VL. The original video is 30 FPS. With our method, only 12 frames could beats 256 frames of Qwen2.5-VL.}
% \label{fig:demo}
% \end{figure}

% \begin{figure}
% \centering
% \includegraphics[width=\linewidth]{figs/demo.png}
% \caption{Two example of our proposed DATE compared with Qwen2.5-VL. The original video is 30 FPS. With our method, only 12 frames could beats 256 frames of Qwen2.5-VL.}
% \label{fig:demo}
% \end{figure}

\begin{table}[]
\centering
\caption{Comparisons with latest methods on LVBench. The baseline is the Qwen2.5-VL-7B model with uniform sampling and their MRoPE.\textbf{ Sampling Strategy:} we compared TASS with AKS (most latest method), and list the computation time for both methods under the same CPU. \textbf{Time Embedding:} We compared our method TIM with timestamps given in prompt.}
\small
\begin{tabular}{c|c|cccc|cc}
\hline
\multirow{2}{*}{Frames}&\multirow{2}{*}{Base}&\multicolumn{4}{c|}{SamplingStrategy}&\multicolumn{2}{c}{Time Embedding}\\\cline{3-8}
&&\multicolumn{2}{c|}{TASS(Ours)}&\multicolumn{2}{c|}{AKS\cite{tang2025adaptive}}&\multicolumn{1}{c|}{TIM(Ours)}&Prompt\\\hline
256&43.7&\textbf{46.7}&\multicolumn{1}{c|}{21.2s}&45.8&21.1s&\multicolumn{1}{c|}{\textbf{44.9}}&42.5\\
128&40.7&\textbf{45.8}&\multicolumn{1}{c|}{6.4s}&44.6&19.2s&\multicolumn{1}{c|}{\textbf{40.2}}&39.4\\
64&38.8&42.6&\multicolumn{1}{c|}{2.7s}&\textbf{43.3}&16.4s&\multicolumn{1}{c|}{\textbf{37.1}}&36.9\\
32&36.8&\textbf{40.9}&\multicolumn{1}{c|}{1.7s}&39.6&13.9s&\multicolumn{1}{c|}{\textbf{37.3}}&35.8\\
16&33.9&\textbf{39.8}&\multicolumn{1}{c|}{1.2s}&33.8&11.7s&\multicolumn{1}{c|}{\textbf{35.7}}&33.1\\\hline
\end{tabular}
\label{tab:frames}
\end{table}

\subsection{Ablation Studies}
We conduct comprehensive ablation studies to evaluate the two core components in DATE: Timestamp Injection Mechanism (TIM) and Temporal-Aware Similarity Sampling (TASS) on Video-MME\cite{fu2024videomme}, LongVideoBench\cite{wu2024longvideobench}, and LVBench\cite{wang2024lvbench}, which are reported in Table~\ref{tab:ablation}.

To further analyze the effectiveness and efficiency of our sampling method, we compare TASS with Adaptive Keyframe Selection (AKS)\cite{tang2025adaptive}, a recent method proposed at CVPR'25, under large range of frame rates (\textbf{from 16 to 256}). As shown in Table\ref{tab:frames}, TASS consistently outperforms AKS across nearly all frame settings, especially at lower frame counts (e.g., +6.0\% at 16 frames), while achieving comparable or even faster sampling times on the same device (Intel Xeon Platinum 8336C (2×32 cores, 2.3 GHz) CPU). These results highlight the efficiency and effectiveness of our sampling design.

Moreover, TIM consistently outperforms the simple "timestamp-in-prompt" method, demonstrating that directly embedding temporal cues into the token space is a more effective way to inject temporal awareness into MLLMs than relying on implicit prompt descriptions.

\subsection{TIM attention analysis}
To investigate the impact of temporal information on video understanding, we visualize attention maps of the baseline and our TIM with timestamp tokens. This experiment is conducted on Question from Fig.\ref{fig:demo}, using 12 input video frames. Since Qwen2.5-vl merges every 2 frames, a total of 6 timestamp tokens are embedded.

As shown in Fig.\ref{fig:att} (left), the baseline exhibits a relatively diffuse attention pattern, indicating that the model relies mainly on content-based similarity across the sequence. In contrast, the attention map of DATE (Fig.\ref{fig:att}, right) reveals a distinct pattern. Notably, video tokens corresponding to the timestamp receive significantly higher attention, suggesting that timestamp tokens act as temporal anchors. They enable the model to associate specific moments with the broader video content.

Furthermore, the explicit temporal cues introduced by timestamp tokens appear to improve the ability to localize frame information. By offering a temporal reference frame for aggregating content across the sequence, the model enhances its contextual understanding of individual video segments.

\subsection{Hyper-parameters Analysis}
\begin{figure}
\centering
\begin{minipage}{0.49\textwidth}
    \centering

    \includegraphics[width=\textwidth]{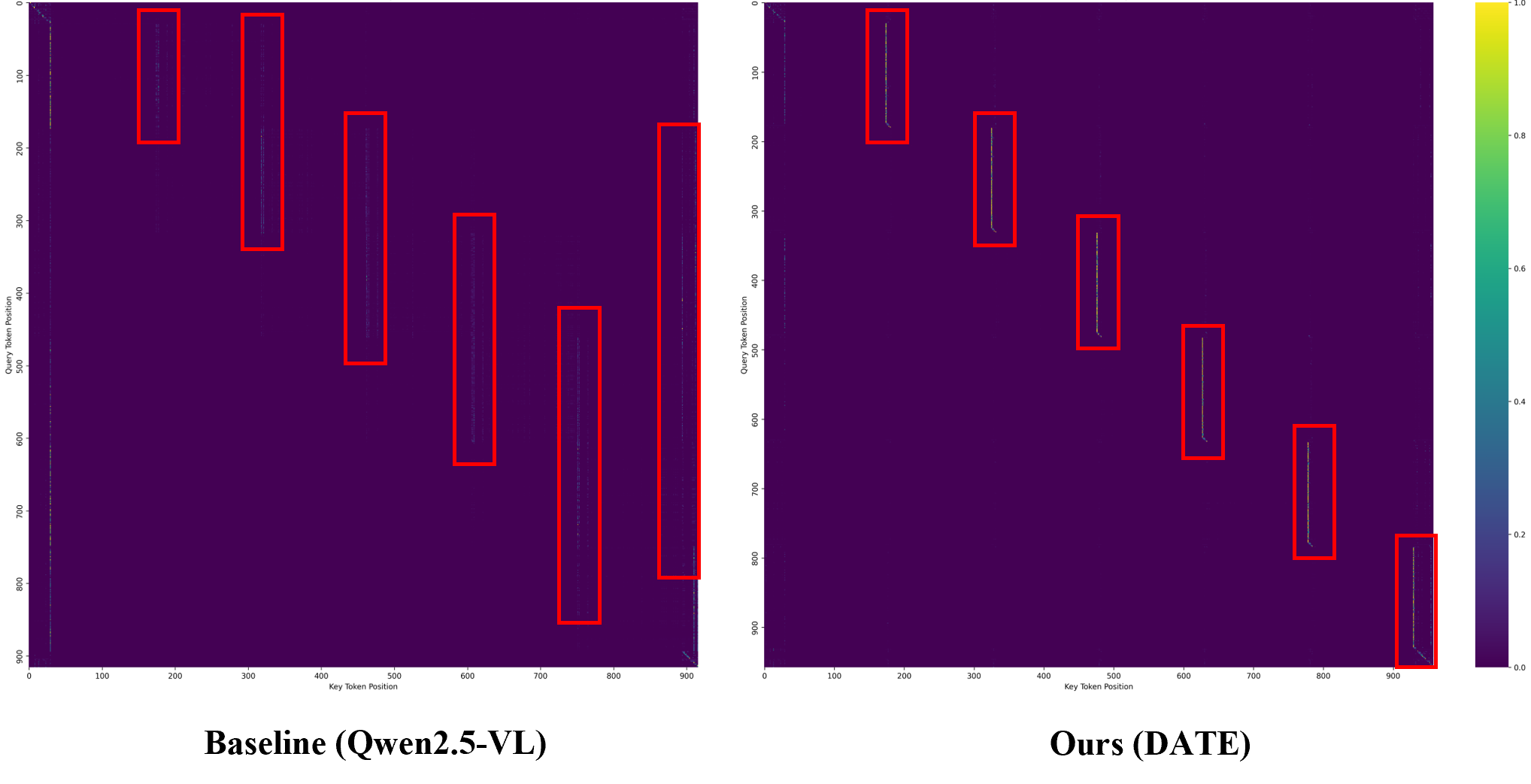}
\caption{Attention maps of our proposed TIM compared with Qwen2.5-VL with 6 times token. Red rectangles label the attention area of each frame's vision tokens. TIM binds timestamps to visual information of the corresponding frame and lead to a scope constraint on attentions.}
\label{fig:att}
\end{minipage}
\hfill
\begin{minipage}{0.49\textwidth}
\centering
    % \includegraphics[width=\textwidth]{figs/event.png}
    % \caption{Comparison of performance related to event-aware tasks in the three benchmarks: Video-MME, LongVideoBench, and LVBench.}
    % \label{fig:event}
    \includegraphics[width=\linewidth]{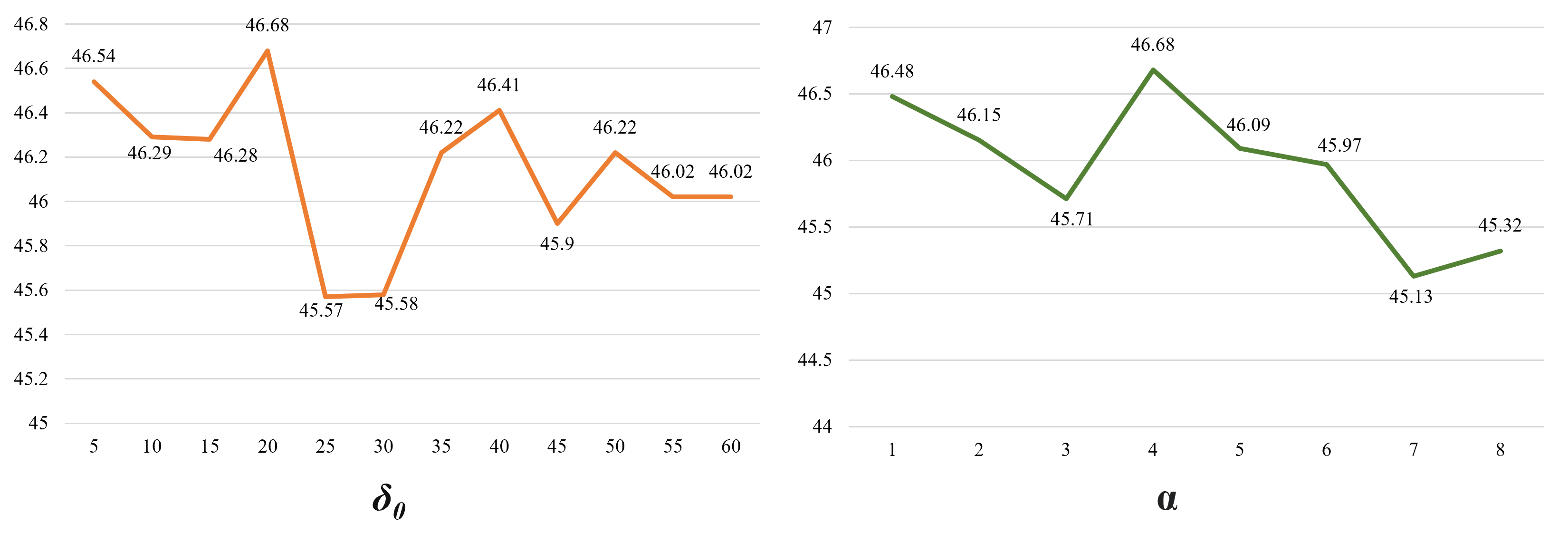}
    \caption{Hyper-parameters analysis of TASS. $\delta_0$ is the initial minimum time interval for sampling, and $\alpha$ controls the candidate sampling frames.}
    \label{fig:param}
\end{minipage}
\end{figure}

As shown in Fig.\ref{fig:param} $\alpha$ controls the number of candidate frames, acting as an effective filtering mechanism to remove distracting information, it achieves the best performance at 4; $\delta_{0}$ constrains the initial temporal range of sampling, demonstrating the stability of the algorithm, which samples well no matter how it is initialized, ensuring continuity between frames and enhancing coverage of key events. Experimental results demonstrate that with appropriate configurations, TASS achieves a good balance between efficiency and temporal awareness.

\section{Conclusion}

In this work, we propose \textsc{DATE}, designed to enhance absolute temporal understanding and event localization in long videos for Multimodal Large Language Models (MLLMs). By timestamp tokens injection mechanism (TIM) and a semantic-driven key event sampling strategy (TASS), our method constructs an explicit and continuous temporal coordinate system without modifying model weights. Extensive experiments on multiple long-video benchmarks demonstrate that \textsc{DATE} significantly improves the model's ability to identify and align over temporally grounded events. Our findings highlight the importance of precise time modeling in long video understanding and open new directions for efficient, inference-time enhancements of pre-trained MLLMs.

% \section*{References}

% {
% \small

% [1] Alexander, J.A.\ \& Mozer, M.C.\ (1995) Template-based algorithms for
% connectionist rule extraction. In G.\ Tesauro, D.S.\ Touretzky and T.K.\ Leen
% (eds.), {\it Advances in Neural Information Processing Systems 7},
% pp.\ 609--616. Cambridge, MA: MIT Press.

% [2] Bower, J.M.\ \& Beeman, D.\ (1995) {\it The Book of GENESIS: Exploring
%   Realistic Neural Models with the GEneral NEural SImulation System.}  New York:
% TELOS/Springer--Verlag.

% [3] Hasselmo, M.E., Schnell, E.\ \& Barkai, E.\ (1995) Dynamics of learning and
% recall at excitatory recurrent synapses and cholinergic modulation in rat
% hippocampal region CA3. {\it Journal of Neuroscience} {\bf 15}(7):5249-5262.
% }

%%%%%%%%%%%%%%%%%%%%%%%%%%%%%%%%%%%%%%%%%%%%%%%%%%%%%%%%%%%%
\newpage
\appendix

\section{Limitations}

Although \textsc{DATE} is an effective approach for enhancing absolute temporal understanding, it still encounters efficiency challenges when dealing with extremely long videos. The reliance on frame-level similarity computation and greedy selection under temporal constraints leads to an inference time that grows approximately linearly with video length. This may result in noticeable latency for hour-long videos—though such delays primarily occur during the initial pass, and subsequent interactions can leverage cached results for near-instant sampling. While reducing the sampling FPS can improve speed, it inevitably compromises precision. Future work may explore more scalable sampling strategies or hierarchical indexing mechanisms to improve runtime efficiency without sacrificing the model’s ability to locate temporally critical events.

\section{TASS Demo}
This is the detail sampling visualization of Fig.\ref{fig:framework}, with 16 sampled \textcolor{red}{red} points and sampling orders labeled.

\begin{figure}[H]
\centering
\includegraphics[width=\linewidth]{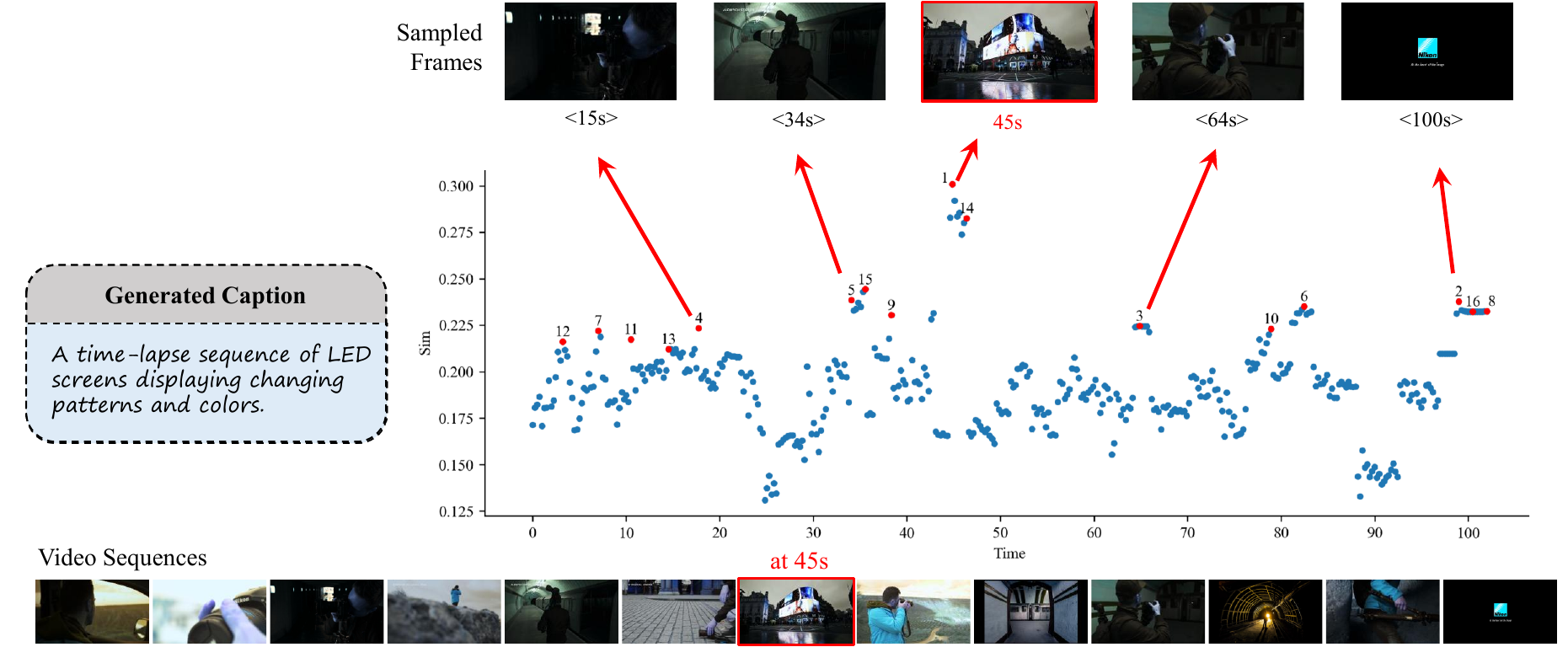}
\caption{Sampling visualization.}
\label{fig:tass_detail}
\end{figure}

\newpage

\section{Qualitative Results and Analysis}
\begin{figure}[H]
\centering
\includegraphics[width=\linewidth]{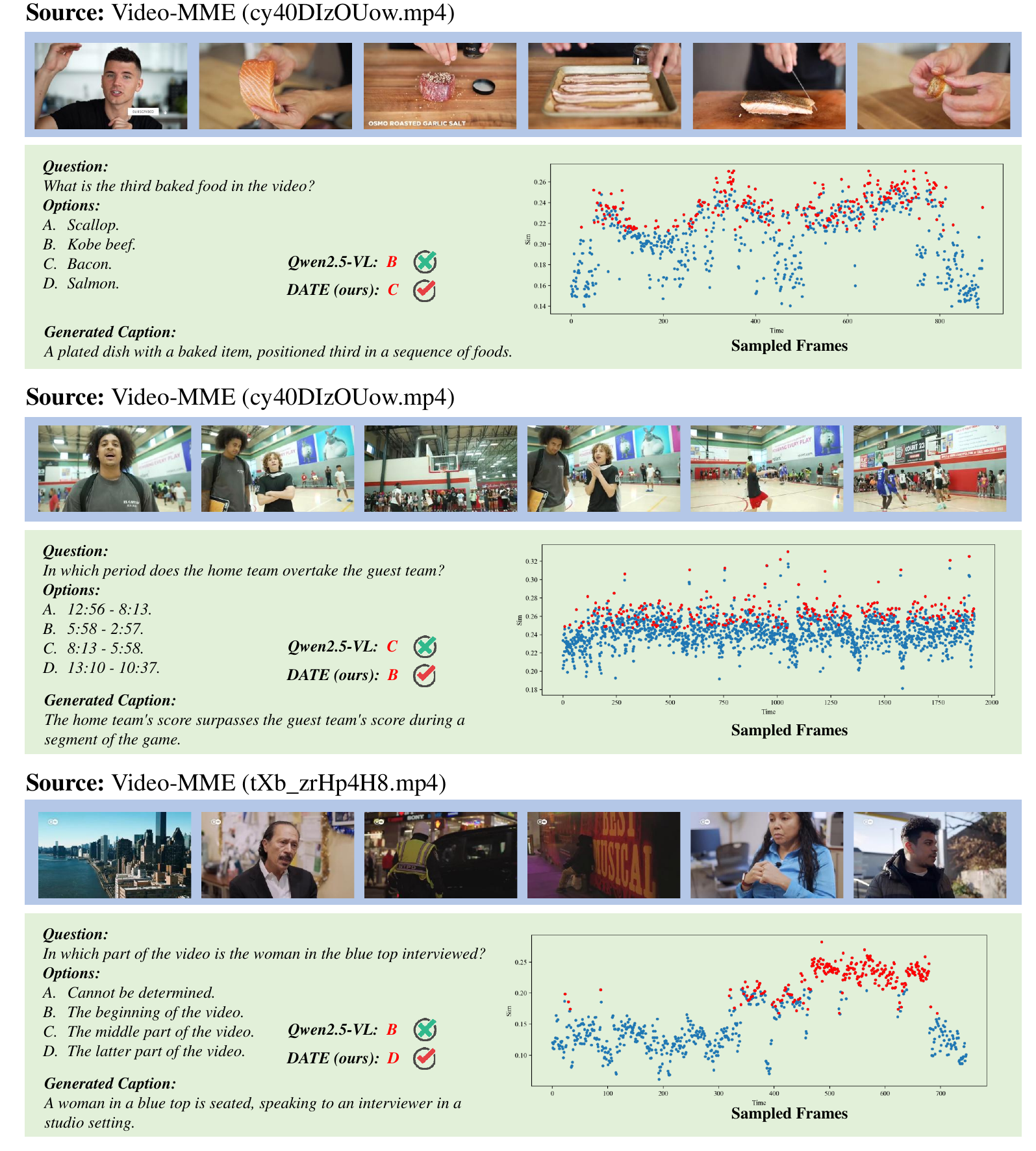}
\caption{Qualitative Results on Video-MME compared with Qwen2.5-VL-7B (1).}
\label{fig:vmme_demo1}
\end{figure}

\begin{figure}[H]
\centering
\includegraphics[width=\linewidth]{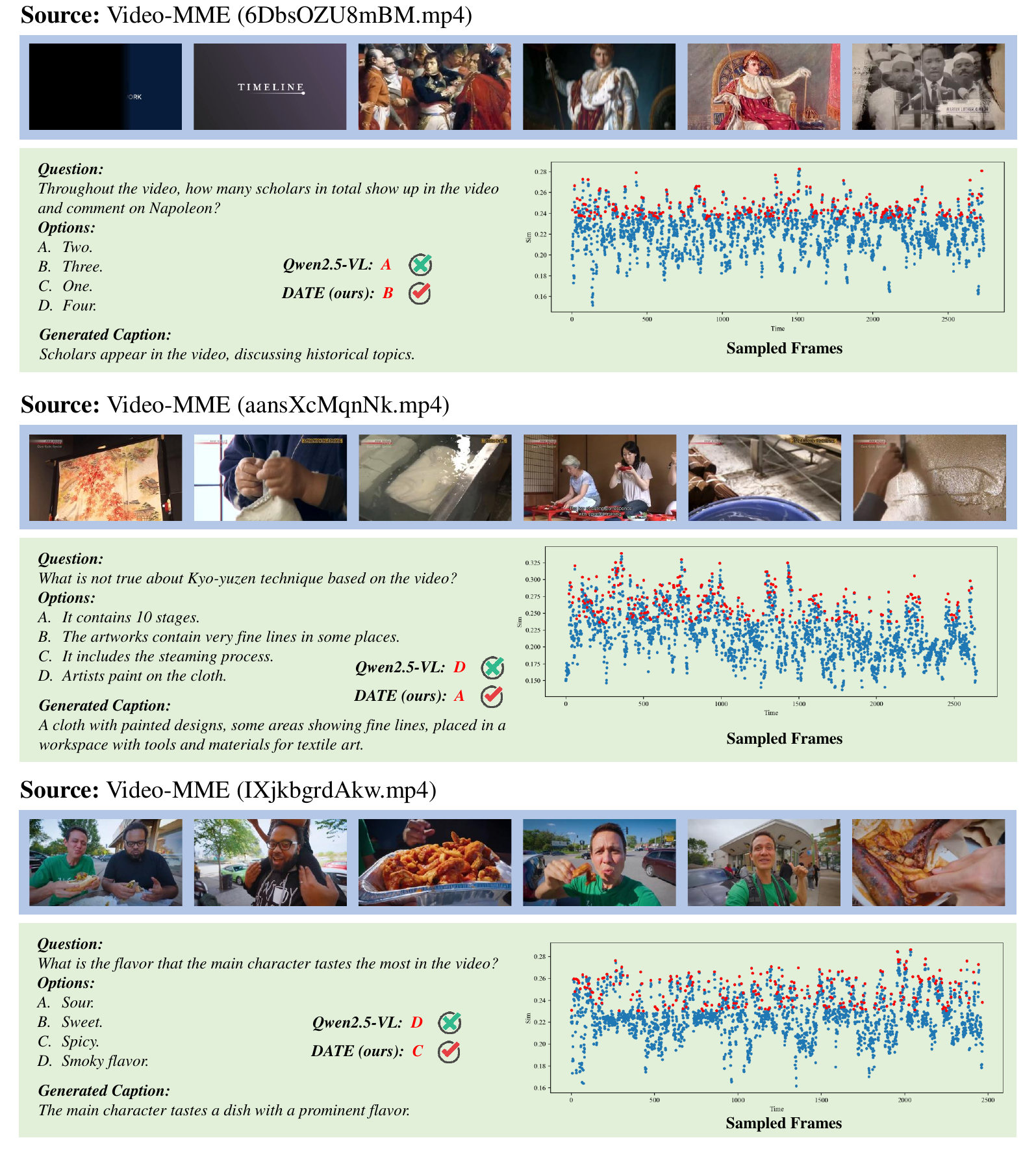}
\caption{Qualitative Results on Video-MME compared with Qwen2.5-VL-7B (2).}
\label{fig:vmme_demo2}
\end{figure}

\begin{figure}[H]
\centering
\includegraphics[width=\linewidth]{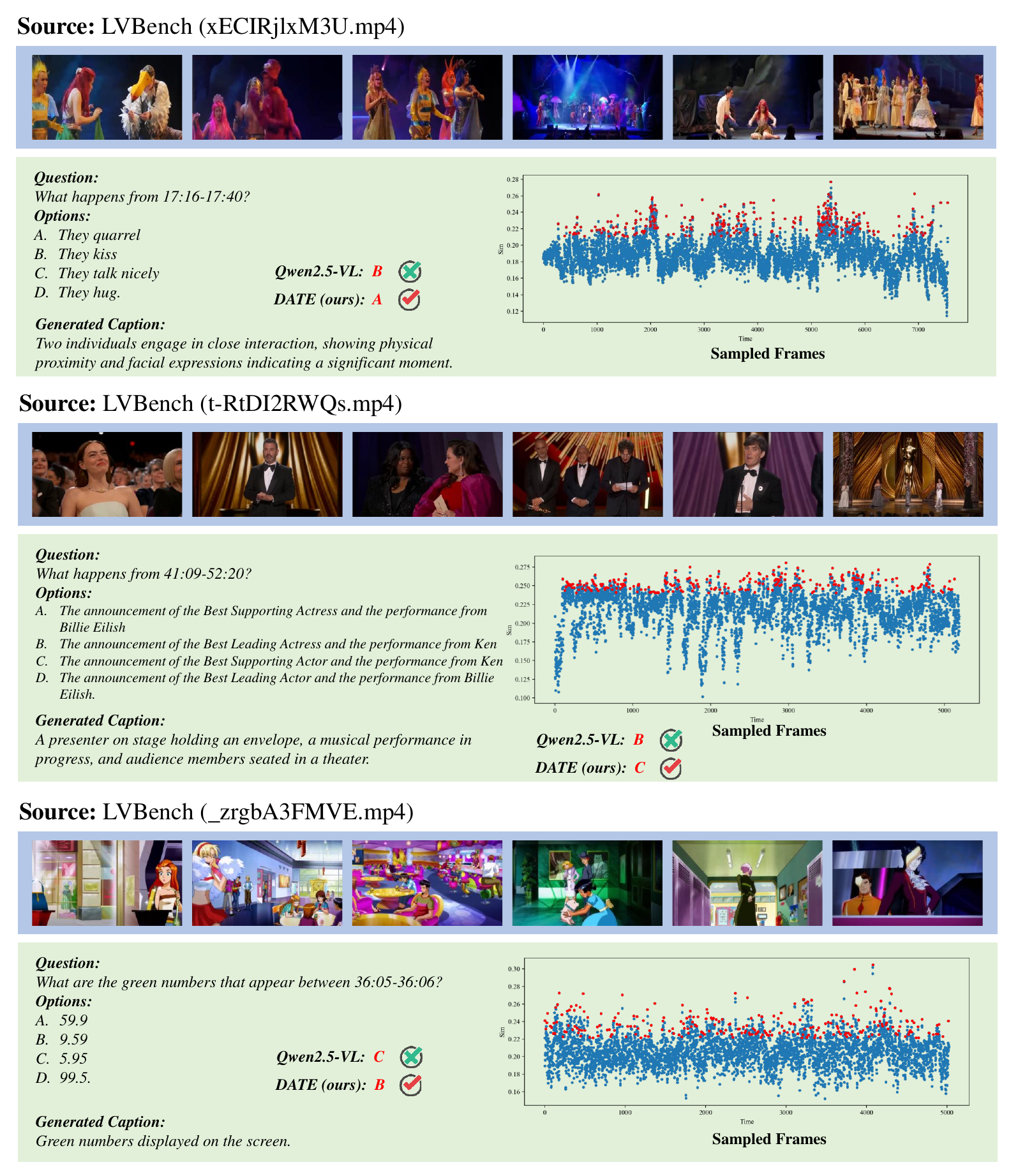}
\caption{Qualitative Results on LVBench compared with Qwen2.5-VL-7B (1).}
\label{fig:lvbench_demo1}
\end{figure}

\begin{figure}[H]
\centering
\includegraphics[width=\linewidth]{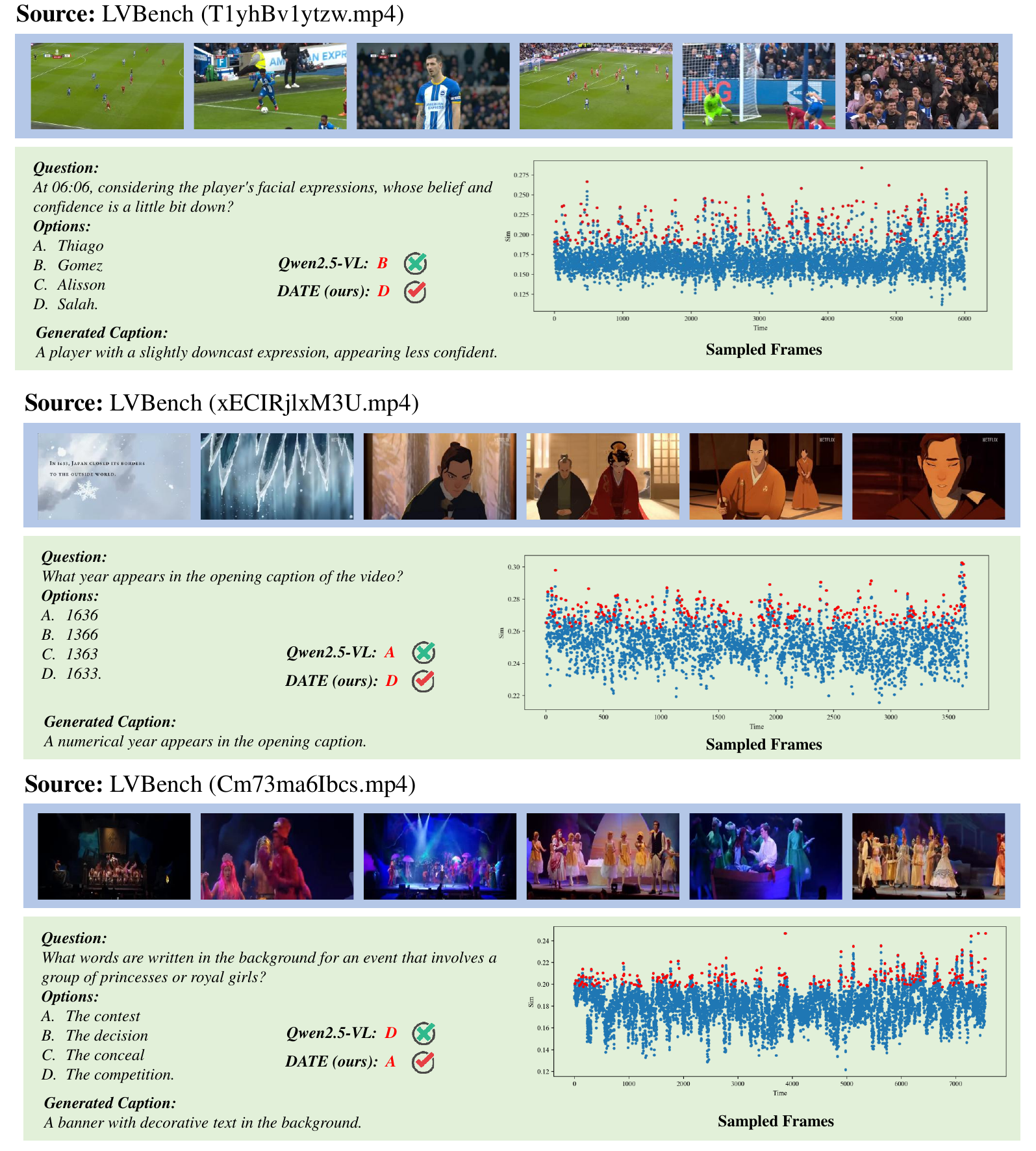}
\caption{Qualitative Results on LVBench compared with Qwen2.5-VL-7B (2).}
\label{fig:lvbench_demo2}
\end{figure}

\begin{figure}[H]
\centering
\includegraphics[width=\linewidth]{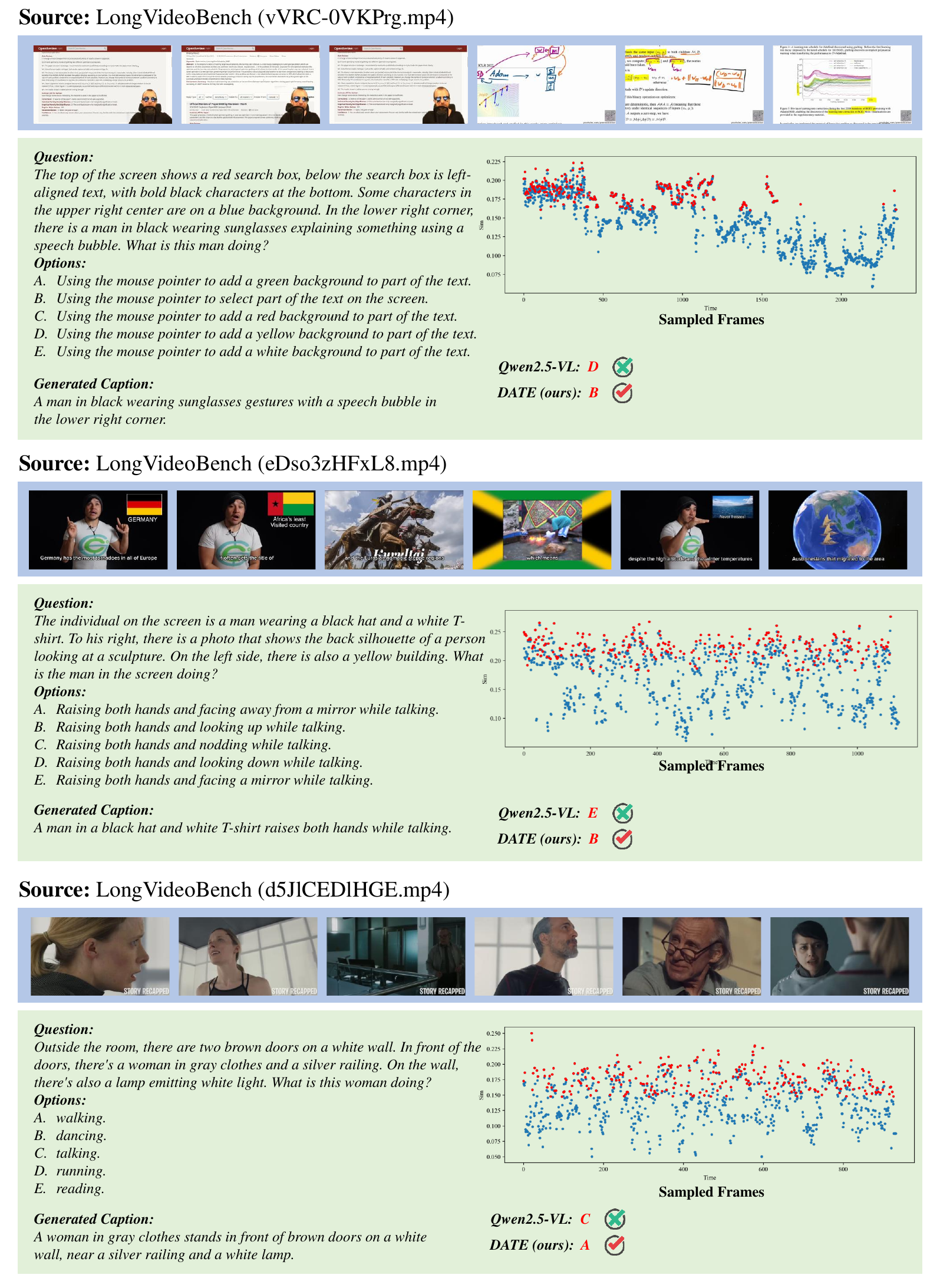}
\caption{Qualitative Results on LongVideoBench compared with Qwen2.5-VL-7B (1).}
\label{fig:longvideobench_demo1}
\end{figure}

\begin{figure}[H]
\centering
\includegraphics[width=\linewidth]{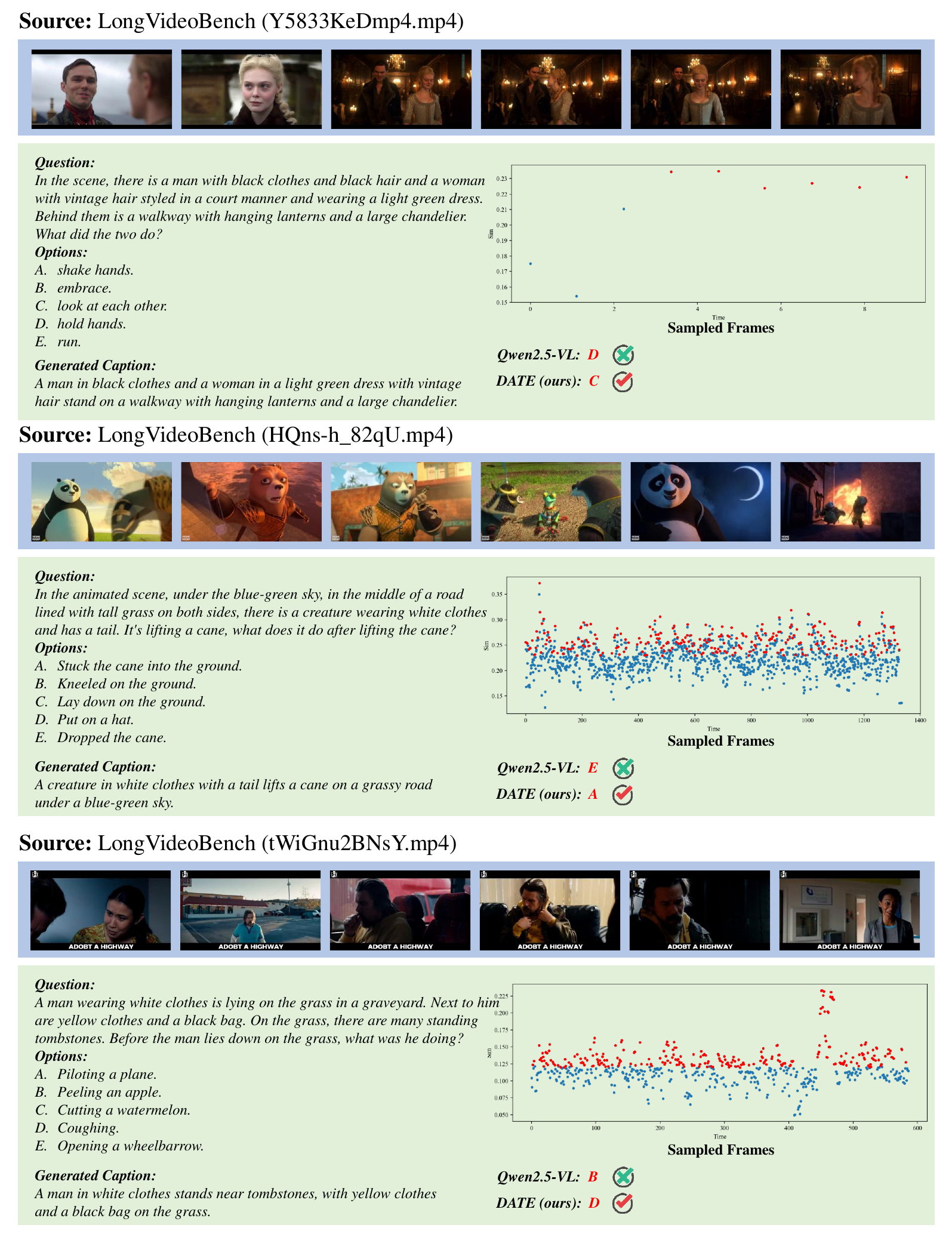}
\caption{Qualitative Results on LongVideoBench compared with Qwen2.5-VL-7B (2).}
\label{fig:longvideobench_demo2}
\end{figure}

We present qualitative results to show the abilities of DATE-7B compared with Qwen2.5-VL-7B across various video understanding benchmarks. Fig.\ref{fig:vmme_demo1},\ref{fig:vmme_demo2},\ref{fig:lvbench_demo1},\ref{fig:lvbench_demo2},\ref{fig:longvideobench_demo1},\ref{fig:longvideobench_demo2} shows qualitative results on Video-MME, LVBench, and LongVideoBench.

\section{Bad Cases}
\begin{figure}[H]
\centering
\includegraphics[width=0.8\linewidth]{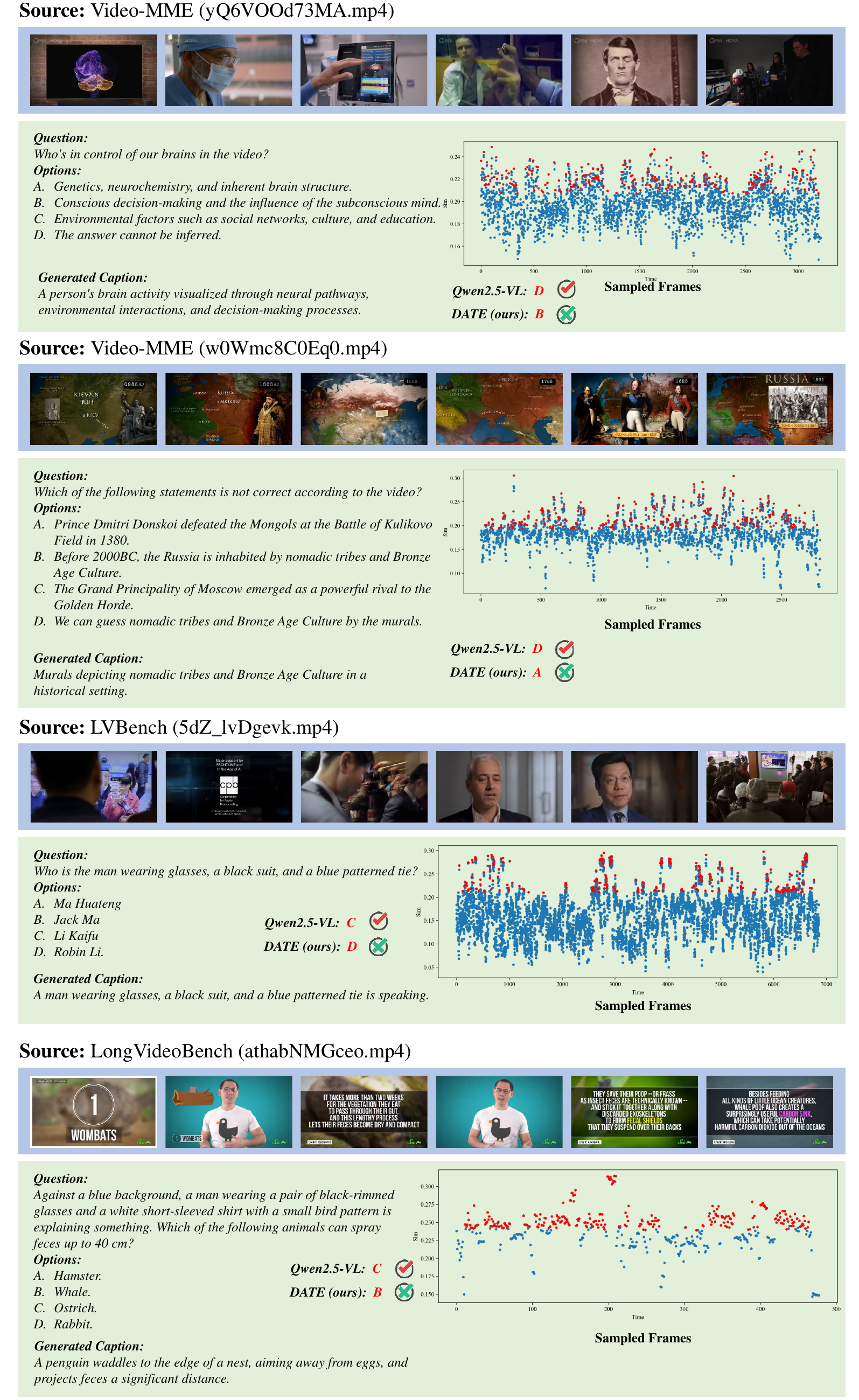}
\caption{Bad cases compared with Qwen2.5-VL-7B.}
\label{fig:badcases}
\end{figure}

While we obtained good boosts across the three benchmarks, we instead made errors compared to the baseline predictions in some cases, as shown in Fig.\ref{fig:badcases}. We believe this may be due to the fact that we introduced additional tokens that increased the processing difficulty of the model, bringing it close to the upper limit of its capacity, thus increasing illusions for certain scenario.

\section{Caption Generation Prompts of TASS}\label{appendix:e}

\begin{tcolorbox}[colback=gray!10!white, colframe=black!75!white, 
  title=Prompt, fonttitle=\bfseries, sharp corners=southwest, 
  boxrule=0.5pt, left=1mm, right=1mm, top=1mm, bottom=1mm]
  
You are an image description assistant. Assume you are currently watching a video, and I will give you a question related to the video. 

Your task is to generate potential image caption based on the question, which is able to find the key image to answer the question.

\vspace{1ex}

Core requirements:

1. The output must be concise, objective, and visually observable facts.

2. Exclude subjective judgments, invisible information, and the specific content the question is asking.

3. Avoid using quantities; use implicit references instead.

4. The question options given are for reference, you can use their commonalities, but not only one of them.

5. Keep the output within 30 words.

\vspace{1ex}

Output format:

Directly output the visual description without any explanations or annotations.

\vspace{1ex}

Here is the question: \{question\}

Output Key Image Caption: 

\end{tcolorbox}

\section{Asset Attribution and License Compliance}

We confirm that all external assets used in this work are properly credited and used in accordance with their licenses. Specifically:

\begin{itemize}
    \item \textbf{Benchmark:} Video-MME (Allows to used for academic research)
    \item \textbf{Benchmark:} LongVideoBench (CC-BY-NC-SA 4.0 license)
    \item \textbf{Benchmark:} LongVideoBench (CC-BY-NC-SA 4.0 license)
    \item \textbf{Model:} Qwen2.5-VL (Apache-2.0 license)
    \item \textbf{Compliance:} No private or proprietary assets were used. All usages comply with academic research standards and ethical guidelines.
\end{itemize}

\bibliographystyle{Ref}  %plainnat,abbrvnat,unsrtnat
\small
\bibliography{Reference}
\normalsize

\end{document}